\def\year{2021}\relax
\documentclass[letterpaper]{article} 
\usepackage{aaai21}  
\usepackage{times}  
\usepackage{helvet} 
\usepackage{courier}  
\usepackage[hyphens]{url}  
\usepackage{graphicx} 
\usepackage{natbib}  
\usepackage{caption} 
\title{Trust Calibration and Trust Respect: A Method for Building Team Cohesion in Human Robot Teams}
\author{
    \\
 Russell Perkins, Zahra Rezaei Khavas, Paul Robinette\\
}
\affiliations{
     \textsuperscript{\rm 1}University of Massachusetts, Lowell\\

    $ 1 $  University Ave, \\
    Lowell, MA 01854\\
    russell\textunderscore perkins@student.uml.edu

}

\begin{document}

\maketitle

\begin{abstract}  
    Recent advances in the areas of human-robot interaction (HRI) and robot autonomy are changing the world. Today robots are used in a variety of applications. People and robots work together in human autonomous teams (HATs) to accomplish tasks that, separately, cannot be easily accomplished. Trust between robots and humans in HATs is vital to task completion and effective team cohesion. For optimal performance and safety of human operators in HRI, human trust should be adjusted to the actual performance and reliability of the robotic system. The cost of poor trust calibration in HRI, is at a minimum, low performance, and at higher levels it causes human injury or critical task failures.  While the role of trust calibration is vital to team cohesion it is also important for a robot to be able to assess whether or not a human is exhibiting signs of mistrust due to some other factor such as anger, distraction or frustration.  In these situations the robot chooses not to calibrate trust, instead the robot chooses to respect trust. The decision to respect trust is determined by the robots knowledge of whether or not a human should trust the robot based on its actions(successes and failures) and its feedback to the human.  We show that the feedback in the form of trust calibration cues(TCCs) can effectively change the trust level in humans.  This information is potentially useful in aiding a robot it its decision to respect trust.
\end{abstract}

\section{Introduction}
    Today, humans regularly interact with intelligent technology thorough devices like Amazon Alexa, smart thermometers, and smart smoke detectors.  With the advent of vacuum cleaner robots, people interact with robots on a daily basis \cite{20}. In addition to household robots, robots have been integrated into first responder and military teams \cite{21}. With this integration not only is there a need for research into how robots can become effective teammates but how they can increase the cohesion and effectiveness of their teams.  The key to being an effective robotic teammate is trust.    
    \par 
    According to Hancock et al. \cite{1} the three classes of factors that influence trust in HRI are, human-related, robot-related and environmental factors. Wagner et al \cite{2} defines trust as \textit{“the reliance by one agent that actions prejudicial to the well-being of that agent will not be undertaken by influential others”}.  Khavas el al. \cite{3}, in an analysis of trust in HRI notes that all definitions of trust incorporate whether robot’s actions and behaviors correspond to human’s interest.  
    \par
     Trust can be broken down into three antecedents, performance, process and purpose \cite{4}. Performance is whether or not the task has been completed, process is how the robot is executing a task and purpose refers to the disposition between the two team members.  When a robot breaks trust it is in one of these three areas.  Breaking trust is referred to as a trust violation \cite{4}. The two types of trust violations are integrity and competence. An integrity violation is when a robot intentionally does something counter to the humans interest.  A competence violation is when a robot incorrectly performs or fails to perform the task. Generally, a human's trust in a robot can be measured by answering the question, "Is the human teammate asking the robot to do tasks that are commensurate with it's capability". Trust calibration is the method in which a robot helps a human to maintain a level of trust that is in line with these capabilities. The goal of calibration is to prevent the humans from ending up in a state of over-trust or under trust. Over-trust is when a person believes that the robot can accomplish something that is outside of its capabilities. Under-trust is when a person does not believe the robot can accomplish tasks for which is is capable \cite{5,7,8}. 
   
    \begin{figure}[t]
\centering
\includegraphics[width=1\columnwidth]{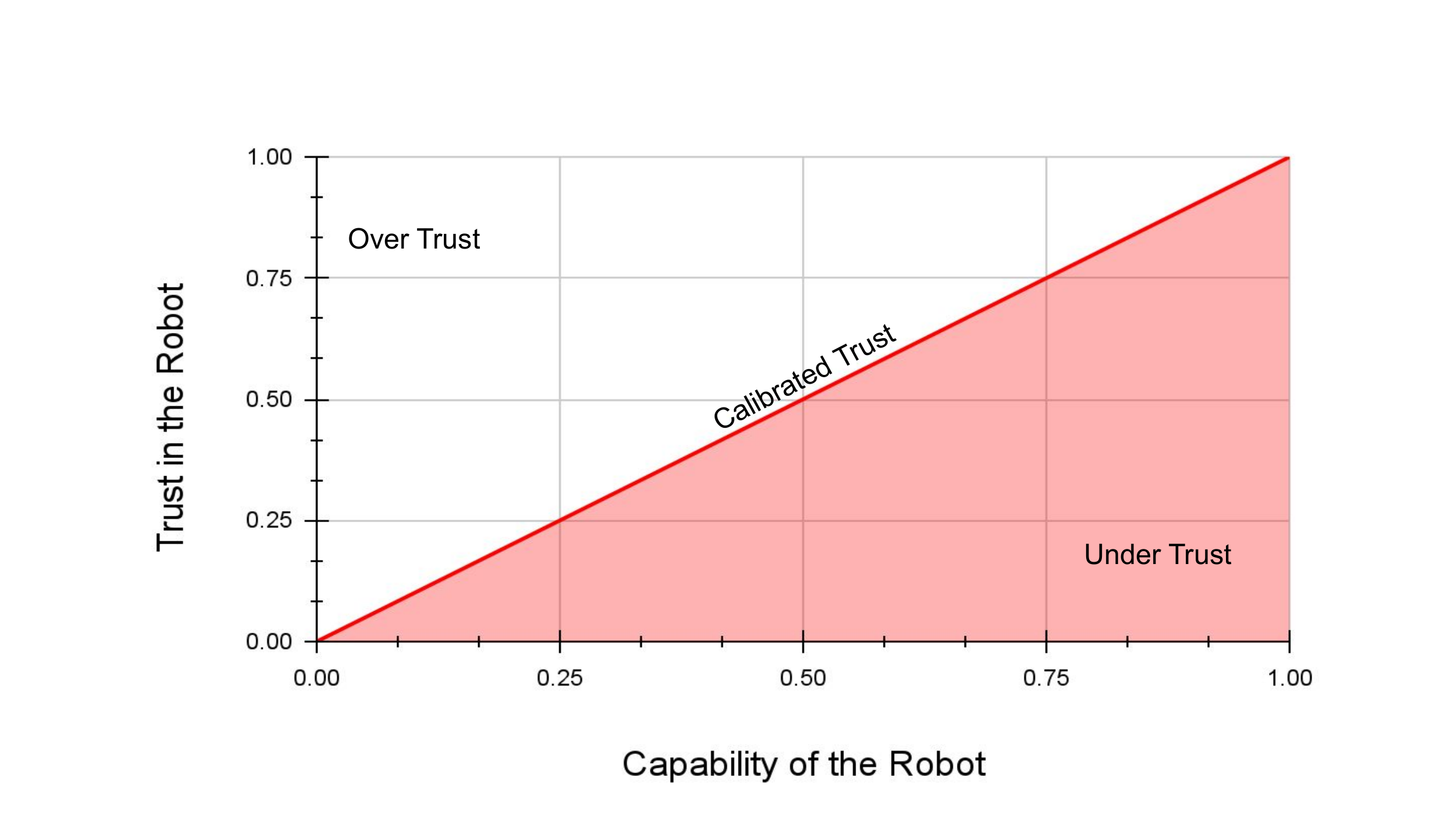} 
\caption{Over trust and under trust as a function of trust in the robot vs. the capability of the robot}.
\label{fig1}
\end{figure}

    \par
    In addition to a robot's ability to calibrate trust, it must also recognise situations that resemble distrust but are, instead, caused by negative emotions. Negative emotions during an interaction with robots are a important area of study in HRI \cite{19}.  Frustrations is a negative emotion that is often mentioned when it comes to human technology interactions \cite{10}. When a human does not trust a robot it can lead to frustration. Humans get frustrated with robots when they have a technical failure. Abd et al. \cite{11} stresses that the level of frustration in a human integrating with a robot was highest when the robot had the largest technical error.  Manuel Giuliani et al. \cite{12} compares the effects of social norm violation and technical failure and shows technical failure cause more frustration.  Weidemann et al. \cite{10}  shows frustration negatively effects the feeling of superiority that the human has over a robot as well as the control that they have over the robot.
\section{Trust Calibration}
     Trust calibration relies of understanding two states, over-trust and under-trust.  The "reliability of the system," is the "probability that a task done by a system will be successful" and is represented by $P_{auto}$.  The user’s estimation of $P_{auto}$ is $P_{trust}$. Over-trust occurs in a situation where $P_{trust}> P_{auto}$. Under-trust occurs when $P_{auto}> P_{trust}$ \cite{13}. The probability of trust is a combination of the successes and failures of the robot and the personality of the human teammate. Similarly frustration is a combination of the failures of the robots and the communication of the robot. Trust manifests itself in the form of trust actions and mistakes manifest themselves in the form of trust violations \cite{13}.   \par
    Trust calibration uses two methods depending on what state of trust the human teammate is in. Figure 1 \cite{4}, shows the relationship between the trust in a robot and the robot's capability. If the person is in a state of over-trust then the robot will attempt to dampen trust.  If the person is in a state of under-trust then the trust repair method is used. These methods are interventions that take the form of trust calibration cues (TCCs). Trust calibration cues can be verbal, visual, audible, physical or a combination of any of these. There is some research concentrated on the effectiveness of different types of TCCs in different situations.  Robinette et al. \cite{8} show a robot that uses anthropomorphic gestures is more effective than having a sign in emergency situations.  Okumura et al. \cite{9} shows that verbal cues are more effective than audible and visual cues.

\section{Respecting Trust}
    In order for a robot to make the decision to respect trust it has to be informed by the environment, its actions and the state of trust that its team mate is in.  There has been work done on predictive models that use the performance of the robot in the form of successes $s$ and failures $f$ to predict the state of trust.   
     Gou et al. \cite{16} tested a beta distribution using an existing data set involving 39 human participants interacting with four drones in a simulated surveillance mission. The proposed method obtained a RMSE of 0.072.  They determined that each person has an optimal set of parameters $\Theta={\alpha_0, \beta_0, w^s , w^f }$ where $w^s$ and $w^f$ are weights applied to the to number of $s$ and $f$, respectively. Theses values are used to determine the parameters of the Beta distribution, $\alpha$ and $\beta$.  The parameter $\Theta$ is updated every time there is a trust decision.  As with the previous methods, the expected value of trust $\hat{t}_i$ at time $i$ is  ($t_i)=\alpha/(\alpha + \beta)$.  Gou and Yang use $\hat{t}_i$ to determine the difference between a the previous trust($\hat{t}_{i-1}$) and $\hat{t}_i$ given a success($p_i =1$) and a failure($p_1=0$).  \par  Gou et al. that it is valid to estimate trust using successes and failures. The use of successes and failures to predict trust is important because it directly ties the the performance of the robot. After a robot has preformed well the trust of its human teammate will increase.  
\subsection{Research Question}    
\textbf{R0:} Can trust calibration actions be used to aid in predicting whether on not a human teammate should trust the robot?
\textbf{H0:}The trust calibration is effective at changing trust regardless of the performance of the robot.  The TCCs can be used as a way to effective estimate future trust.

\section{Experimental design}
    Our experiment is an online game conducted with participants from Mechanical Turk.  The game is a simulation of a search and rescue (S\&R) scenario where a human and a robot, referred to as agents, move through a simulated building to identify simulated victims, referred to as targets.  There are two types of targets, gold stars which are worth a $100$ points and red circles which are worth $-100$ points.  Each agent has a predefined area of discovery.  When each of the agents find a target, they need to take an action to select the target. Selecting targets is optional for the human teammate, however, selecting only the gold stars will maximize the team score. The game consists of 10 rounds. At the end of each round the human teammate makes a trust action. The trust action is a blind decision whether to integrate or discard the map and targets discovered by the robot.  After the decision the collective team score is updated.  Because it is a blind decision the human is forced to make their decision based on robots prior performance. When the robots information is integrated all the scores gained by the robot is added to the overall score of the team and the areas searched by robot cannot be searched again by the human.  If the human agent discards the information of the robot the robot score will not be added to the team score and the areas search by robot can be searched by human later.  After making trust decision, the human will be able to see the score and targets gained by robot in this round. Table 2 shows the targets score and TCC used in each round of the game.  Three manipulation questions were used each asking specific questions regarding to the mechanics of the survey. If a participant failed two out of three of these manipulation questions they were excluded from the survey.

    \begin{table}[]
    \centering
    \scriptsize
    \begin{tabular}{ |p{3cm}|p{3cm} | }
    
     \hline
   All Negative TCCs & All Positive TCCs\\
     \hline
     \hline
     I am sorry, I was having difficulty identifying the correct target.  I will do better next round.    &I am not going to be able to accurately identify targets next round.\\
     \hline
    I am sorry, I am still having trouble with identification.  Let me try something different to see if that will help.& I am still having trouble identifying targets.  \\
     \hline
     
     \end{tabular}
    \caption{Trust calibration cues used.}
    \label{table1}
    \end{table}

\begin{table}[]
    \centering
 \scriptsize
\begin{tabular}{|l|l|l|l|l|}
\hline
\begin{tabular}[c]{@{}l@{}}Round\\ number\end{tabular} & \begin{tabular}[c]{@{}l@{}}Gold \\ stars\end{tabular} & \begin{tabular}[c]{@{}l@{}}Red\\ circles\end{tabular} & Score\\ \hline
1                                                      & 2                                                     & 3                                                     & -100  \\ \hline
2                                                      & 1                                                     & 4                                                     & -300    \\ \hline
3                                                      & 1                                                     & 2                                                     & -100   \\ \hline
4                                                      & 2                                                     & 3                                                     & -100        \\ \hline
5                                                      & 0                                                     & 2                                                     & -200        \\ \hline
6                                                      & 0                                                     & 1                                                     & -100        \\ \hline
7                                                      & 0                                                     & 1                                                     & -100        \\ \hline
8                                                      & 0                                                     & 2                                                     & -200        \\ \hline
9                                                      & 2                                                     & 3                                                     & -100        \\ \hline
10                                                     & 1                                                     & 2                                                     & -100        \\ \hline
\end{tabular}
    \caption{{Number of targets and the scores for the all negative surveys}}
    \label{tab:my_label}
\end{table}

\begin{table}[]
    \centering
 \scriptsize
\begin{tabular}{|l|l|l|l|l|}
\hline
\begin{tabular}[c]{@{}l@{}}Round\\ number\end{tabular} & \begin{tabular}[c]{@{}l@{}}Gold \\ stars\end{tabular} & \begin{tabular}[c]{@{}l@{}}Red\\ circles\end{tabular} & Score   \\ \hline
1                                                      & 3                                                     & 2                                                     & 100     \\ \hline
2                                                      & 1                                                     & 0                                                     & 100    \\ \hline
3                                                      & 2                                                     & 0                                                     & 200   \\ \hline
4                                                      & 4                                                     & 1                                                    & 300        \\ \hline
5                                                      & 4                                                     & 0                                                    & 400       \\ \hline
6                                                      & 4                                                     & 3                                                     & 100        \\ \hline
7                                                      & 1                                                     & 0                                                     & 100       \\ \hline
8                                                      & 2                                                     & 0                                                     & 200        \\ \hline
9                                                      & 3                                                     & 2                                                     & 100        \\ \hline
10                                                     & 4                                                     & 3                                                     & 100        \\ \hline
\end{tabular}
    \caption{Number of targets and the scores for the all positive surveys}
    \label{tab:my_label}
\end{table}
\section{Preliminary results}
Four surveys were conducted using participants from Amazon Mechanical Turk.  Two of the surveys were controls meaning that they had no TCCs. One of the controls had all negative scores and one of the controls had all positive scores.  The other two surveys had TCCs located after the third round and on every subsequent round for 10 rounds.  Table 2 is a list of the scores for the all negative surveys that were conducted.  The all positive trust calibration survey as compared to the all positive control and likewise for the all negative versions.  Table 3 is a list of the scores for the all positive surveys that were conducted. Because each survey had a different number of participants after the data was filtered for the manipulation questions the overall trust score is given in a percentage calculated using Equation 1.
\begin{equation}
    \frac{\mbox{number of participants who integrated}}{\mbox{total number of participants}} 
\end{equation}
The all positive TCC survey had a drop from 23 participants to 13 participants after the first calibration. The all negative TCC survey showed an increase from 10 to 18 participants after the calibration.  Figure 2 is a plot of the trust scores for the all positive surveys and figure 3 is a plot of the trust scores for the all negative surveys.  When the two trust calibration surveys were compared with each other it showed that it took two rounds for the participants to ignore the feedback from the robot regardless of the goal of the TCC.  

    \begin{figure}[t]
\centering
\includegraphics[width=1\columnwidth]{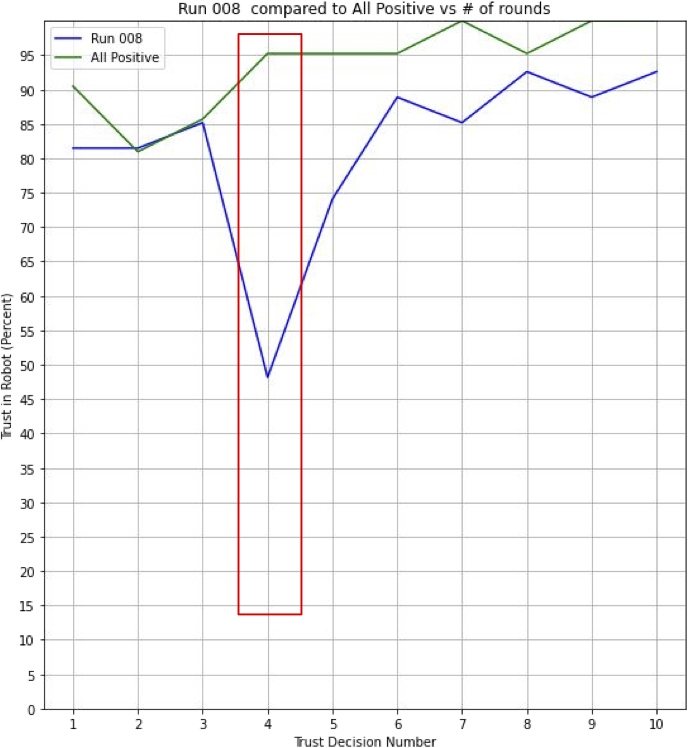}
\caption{Plot of function of percent trust vs. number of rounds of the all positive surveys.  The blue line is the TCC test group and the green line is the control group}. The red box shows the round after the TCCs began.
\label{fig1}
\end{figure}
    \begin{figure}[t]
\centering
\includegraphics[width=1\columnwidth]{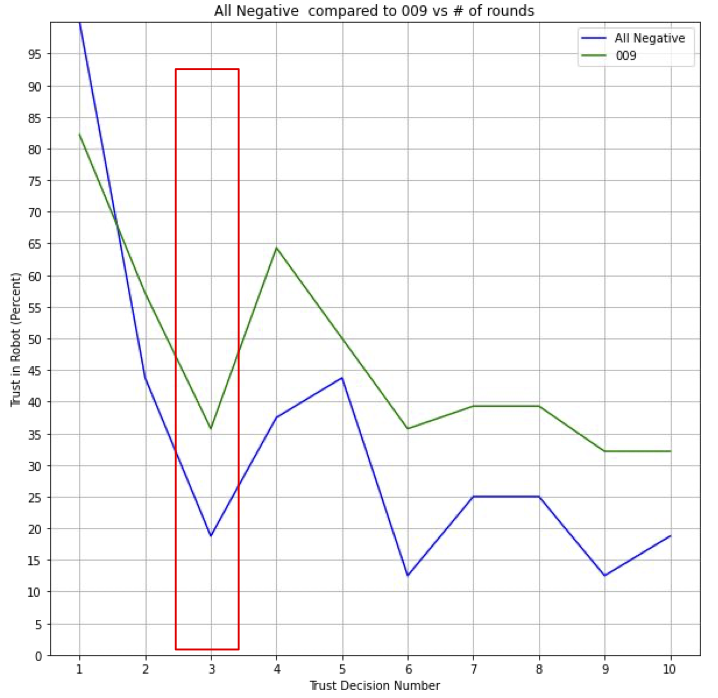}
\caption{Plot of function of percent trust vs. number of rounds of the all negative surveys.  The green line is the TCC group and the blue line is the control group}.  The red box shows the round after the TCCs began.
\label{fig1}
\end{figure}
\section{Conclusion and Future Work}
    Human robot teams are no longer the realm of fiction. In order to develop effective teams there must be a framework so that humans an robots can work together as a cohesive unit. The cornerstone of this framework is going to be the ability for the robot to communicate what its capability is and inform its human counterpart when their trust is not in line with its capabilities. If a robot can identify when its human teammate is showing signs of distrust for external reason (i.e. frustration) it can chose to act on its own to aid its teammate. Frustration and cognitive load has been shown to mirror the effects of distrust resulting in under-trust \cite{19}. A robot that can recognize when frustration presents as distrust will be able to help its teammate by not adding to cognitive load.  
    \par
    Trust calibration cues have reliably can shift trust in the direction they are designed for.  Both of the TCC experiments showed that the feedback that the robot provided corresponded with the desired change in trust.  When we attempted to dampen trust we effectively lowered the trust value despite the good performance.  Likewise, when we repaired trust, there was an increase in the trust score despite the low performance of the robot.  This is important for trust estimation because, if, in a prior round the robot attempted to repair trust then it should estimate that its human teammate will have a higher level of trust.  If in fact they do not, potentially, the robot could make the decision to respect the trust state of the teammate because there is an external force causing an artificial trust state.
    \par 
    We are a long way from having a reliable prediction model that will aid a robot in choosing to respect trust or calibrate trust.  We have shown that TCCs are effective at changing trust in the desired direction which will aid in the model.  As mentioned in the beginning of this paper, different trust violations respond differently to different trust calibrations.  We are going to conduct an experiment to see what that difference is.  We also have to figure out a way to combine performance and calibration into a predictive model.  After the basic model has been created in the future we would like to add to it to include biological signals such as heartbeat. Ultimately we want to test the model with people where an external force, like frustration, is applied to see if a robot, using our model, can accurate determine when to respect trust or calibrate trust.
\bibliography{ref.bib} 

\end{document}